\pdfoutput=1

\documentclass[11pt]{article}

\usepackage[final]{acl}

\usepackage{times}
\usepackage{latexsym}
\usepackage{booktabs}
\usepackage{multirow}
\usepackage{tabularx}
\usepackage{amssymb}
\usepackage{amsmath}
\usepackage[T1]{fontenc}

\usepackage[utf8]{inputenc}
\usepackage{enumitem}
\usepackage{microtype}

\usepackage{inconsolata}

\usepackage{graphicx}
\usepackage{mdframed}
\usepackage{soul}

\title{FiMMIA: scaling semantic perturbation-based membership inference across modalities}

\author{Anton Emelyanov \\
  SberAI \\
  \texttt{login-const@mail.ru} \\\And
  Sergei Kudriashov \\
  Sber, HSE University\\
  \texttt{sakudryashov@hse.ru} \\\\\And
  Alena Fenogenova \\
  SberAI\\
  \texttt{alenush93@gmail.com} \\
}

\begin{document}
\maketitle
\begin{abstract}

Membership Inference Attacks (MIAs) aim to determine whether a specific data point was included in the training set of a target model. Although there are have been numerous methods developed for detecting data contamination in large language models (LLMs), their performance on multimodal LLMs (MLLMs) falls short due to the instabilities introduced through multimodal component adaptation and possible distribution shifts across multiple inputs. In this work, we investigate multimodal membership inference and address two issues: first, by identifying distribution shifts in the existing datasets, and second, by releasing an extended baseline pipeline to detect them. We also generalize the perturbation-based membership inference methods to MLLMs and release \textbf{FiMMIA} --- a modular \textbf{F}ramework for \textbf{M}ultimodal \textbf{MIA}.\footnote{The source code and framework have been made publicly available under the MIT license via \href{https://github.com/ai-forever/data_leakage_detect}{link}.The video demonstration is available on \href{https://youtu.be/a9L4-H80aSg}{YouTube}.}
Our approach trains a neural network to analyze the target model's behavior on perturbed inputs, capturing distributional differences between members and non-members. Comprehensive evaluations on various fine-tuned multimodal models demonstrate the effectiveness of our perturbation-based membership inference attacks in multimodal domains.

\end{abstract}

\section{Introduction}
The development of MLLMs has exceeded expectations \cite{Liu2023a, Lin2023VILAOP}, showcasing extraordinary performance on various multimodal benchmarks \cite{mera, Lu2022LearnTE, Liu2023MMBENCIY, Song2024MilebenchBM}, even surpassing human performance. However, due to the partial obscurity associated with MLLMs training or fine-tuning \cite{GPT4TechnicalReport, Reid2024GeminiU}, it remains challenging to definitively ascertain the impact of training data on model performance, despite some works showing the employment of the training set of certain datasets \cite{Liu2023a, Chen2023InternVLSC, Bai2023QwenVA}. The issue of data contamination occurs when training or test data of benchmarks is exposed during the model training or fine-tuning phase \cite{Xu2024BenchmarkingBL} and could potentially instigate inequitable performance comparisons among models.

Although numerous works in the field of LLMs have proposed methods for detecting data contamination \cite{smia, Hu2022MembershipIA, song-etal-2025-text, NEURIPS2024_b2c89231}, MLLMs, due to their various modalities that, in most implementations, lack corresponding target tokens for multimodal inputs, while multiple training phases, common for MLLM training, complicate an inference when one tries to apply these methods directly. Therefore, there is a necessity in a multimodal contamination detection framework specifically tailored for MLLMs.
Our main contributions can be summarized as follows:
\begin{itemize}[nosep]
    \item We extended the work of \citet{das2024blind} to multimodal data and assessed image as well as recent text MIA benchmarks 
    \cite{fu2024wikimia24, hallinan2025wikimiahard} for distribution shifts and find 
    that even the \textit{most recent proposed benchmarks are subject to distribution shifts between member and non-member data}.
    \item We \textit{release an attack pipeline for image, video and audio data}, that collects various statistics from the dataset distribution and trains a classifier on top to distinguish members from non-members without any signal from the target model. 
    \item We \textit{extend perturbation-based MIA methods to MLLMs}, revealing their effectiveness and transferability even at scale with billion-parameter models.
    \item We \textit{release a modular framework FiMMIA} supporting diverse datasets, modalities, and neighbor generation methods.
\end{itemize}

\section{Related Work}
\label{sec:related_work}

\subsection{Data contamination and distribution shifts hinder reliable evaluations}
\label{subsec:contamination&shifts}
Preserving training data confidentiality is critical for LLMs, as their datasets can contain sensitive private information and tests \cite{62, 24}. Additionally, data contamination between training and test sets undermines benchmark reliability and complicates model comparison \cite{balloccu-etal-2024-leak, sainz-etal-2023-nlp}, driving recent adoption of dynamically updated benchmarks \cite{livebench}.


Distribution shifts pose significant risks as neural networks' ability to extract subtle correlations makes them vulnerable to adversarial examples \cite{moayeri2022advnattradeoff}, spurious correlations in explanations \cite{ribeiro2016lime}, and data poisoning \cite{souly2025poisoning}.
Recent studies have also found that modern LLMs are capable of intensional \textit{sandbagging}, i.e., strategically underperforming during the evaluations in the presence of an incentive to do so \cite{weij2024ai}. In other words, capable LLMs can intensionally manipulate their logprobs, which poses an additional challenge both for capability elicitation and loss-based MIA attacks~\footnote{Such behavior is only possible if the evaluation data or environment presents enough evidence to distinguish it from the training environment, even due to subtle cues.}.

\subsection{Membership inference attacks aim to solve the problem}
\label{subsec:membershipinferencedesign}

Membership Inference Attacks (MIAs) determine whether a data sample was part of a model's training set \cite{53} or originates from the general distribution. As noted by \cite{carlini2022membershipinferenceattacksprinciples}, this constitutes a hypothesis testing task that crucially relies on the i.i.d. assumption.

Membership Inference Attacks have been the subject of considerable research across a variety of machine learning models, including classification models \cite{38, 58, 10}, generative models \cite{20, 22, 7}, and embedding models \cite{54, 40}. The appearance of LLMs has likewise led to numerous studies investigating membership inference attacks against them  \cite{42, 17, 52, 41}. However, the field of MIAs for multimodal models is still in its nascent stages and requires further exploration, facing challenges due to the absence of targets for modality-related tokens, instabilities from multimodal adaptation etc. Several methods \cite{30, 25} proposed to conduct MIAs based on the similarity between an image and its associated text label. However, this technique is limited to the presence of  a paired entry (pair image/text), not the presence of a solitary image or text sequence.

MIAs are commonly categorized into metric-based and shadow model-based approaches \cite{24}. Metric-based MIAs \cite{62, 48, 57, 52} compare model output statistics against a threshold, while shadow model-based methods \cite{53, 48} require computationally expensive model replication. Recent work has introduced semantic MIAs \cite{koike2025machinetextdetectorsmembership, smia} that exploit local model properties through sample perturbations. We extend this semantic approach to image, audio, video, and text modalities.

\begin{figure*}
    \centering
    \includegraphics[width=\linewidth]{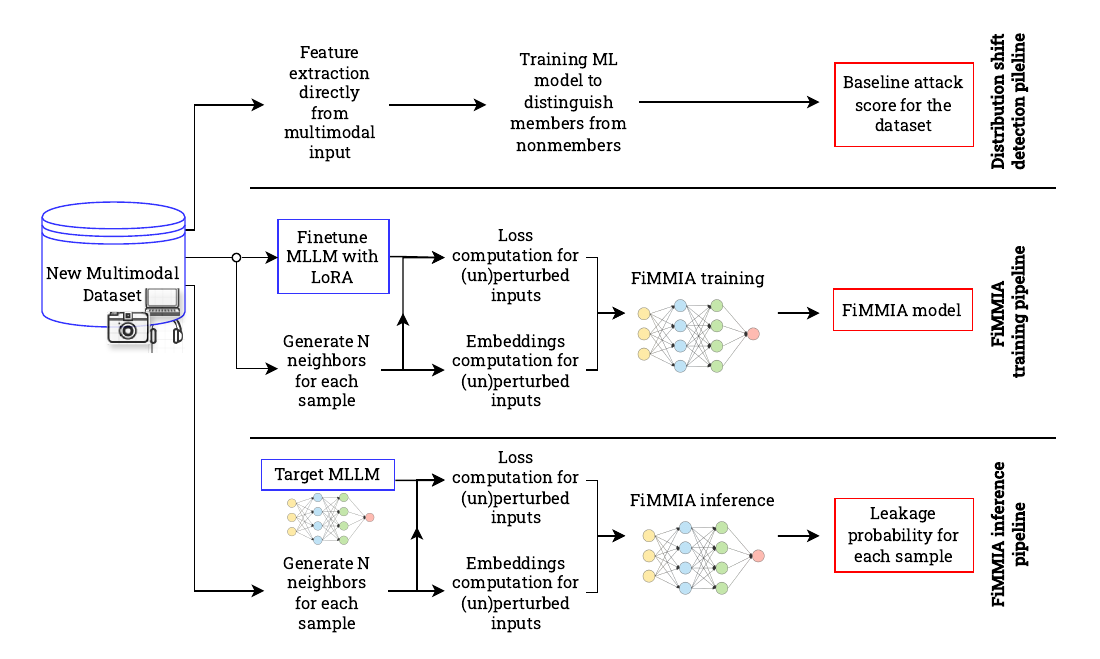}
    \caption{Overview of FiMMIA Inference pipeline for MLLMs.
Inputs to the pipeline are shown \textcolor{blue}{in blue}. Outputs of the pipeline are shown \textcolor{red}{in red}.}
    \label{fig:overview}
\end{figure*}

\begin{table}[t]
\centering
\scriptsize
\begingroup
\setlength{\tabcolsep}{4pt}
\begin{tabular}{p{0.1cm} l l c} 
\toprule
\textbf{} & \textbf{Dataset / task} & \textbf{Best reported(\%)} & \textbf{Our baseline(\%)} \\
\midrule
\multirow{4}{*}{\rotatebox[origin=c]{90}{\textbf{text}}}
& WikiMIA-hard & 64.0 \tiny{\cite{hallinan2025wikimiahard}} & $57.7 \pm 2.5$  \\
& WikiMIA-24 & 99.8 \tiny{\cite{fu2024wikimia24}} & $ 99.9 \pm 0.1$  \\
& VL-MIA-Text (32 tok.) & 96.2 \tiny{\cite{li2024vlmmembership}} & $84.9 \pm 4.0$  \\
& VL-MIA-Text (64 tok.) & 99.3 \tiny{\cite{li2024vlmmembership}} & $95.5 \pm 0.9$ \\
\midrule
\multirow{5}{*}{\rotatebox[origin=c]{90}{\textbf{image}}}
& VL-MIA-Flickr & 94.2 \tiny{\cite{yin2025black0box}} & $99.1 \pm 0.4$  \\
& VL-MIA-Flickr-2k & 74.0 \tiny{\cite{li2024vlmmembership}} & $98.6 \pm 0.4$  \\
& VL-MIA-Flickr-10k & NA & $99.3 \pm 0.1$  \\
& VL-MIA-DALL-E & 84.0 \tiny{\cite{yin2025black0box}} & $99.9 \pm 0.1$  \\
& LAION-MI* & 2.42 \tiny{\cite{dubiński2023realistic}}& $1.11 \pm 0.1$  \\
\bottomrule
\end{tabular}
\endgroup
\caption{AUC-ROC Evaluations of image and text MIA datasets for the occurrence of distribution shifts between members and non-members data.
* corresponds to TPR@1FPR instead.}
\label{tab:distributionshift}
\end{table}

\vspace{-0.5em}
\section{FiMMIA}
\label{sec:FiMMIA}
\vspace{-0.5em}
\subsection{Overview}
The system is the first collection of models and pipelines for membership inference attacks against LLMs, built  and evaluated initially on the Russian
language, and extendable to any other language or dataset that matches the format. 
The pipeline supports different modalities: image, audio and video and is fully open source\footnote{\url{https://github.com/ai-forever/data_leakage_detect}} We also provide pretrained FiMMIA models\footnote{\url{https://huggingface.co/collections/ai-forever/fimmia}}. Although in our experiments we focus on MERA datasets \cite{mera} to ensure independence in the split between members and nonmembers, the presented pipeline is built with the idea of supporting modular extension and is intended to be easily adopted. 
Overall, the system is a set of models and Python scripts in a GitHub repository that supports three major functionalities: 1) a baseline attack based on distribution statistics, intended to ensure the reliability of multimodal MIA methods; 2) inference scripts for the FiMMIA model; 3) a training pipeline for new datasets. The main system parts are shown at~\autoref{fig:overview}, the general pipeline for multimodal MIA is described in~\autoref{sec:method}.


\subsection{Multimodal membership inference benchmarks suffer from distribution shifts}
\label{subsec:multimodaldistributionshift}

Recently, \cite{das2024blind} have evaluated common textual membership inference benchmarks using blind statistical methods, and have found that they suffer from  distribution shifts, with their baseline methods outperforming best membership inference attacks on these datasets. An introduction of embedding model into the pipeline \cite{smia, hu2022m04i0} obviously makes the matter even worse, as they shine in tasks related to the separation of different distributions. This fact has, e.g. been recently utilized by \cite{miyamoto2025openlvlm0mia}, who have also acknowledged the problem, and used a DINO-V2 \cite{oquab2023dinov20} to extract image features and showed that VL-MIA member and non-member data suffer from a distributional mismatched introduced by the generative nature of non-member samples with AUC-ROC of 94.9\% using their method. 
There are reasons for us to argue against this approach. Foremost, the usage of advanced deep learning model still poses some of the threats outlined above. Thus, we extend the work of \cite{das2024blind} to multimodal data and, to our surprise, find that attacks that directly use features obtained from the dataset samples in absence of any information from the target model outperform best known attacks on most multimodal MIA benchmarks. 

\subsection{Distribution shift detection \& baseline attacks}

Essentially, for each input sample from the dataset with specified members and non-members we extract common heuristic (e.g. SIFT, LBP histogram) or spectral features, and them as inputs to a shallow ML model (e.g. logistic regression or gradient boosting)\footnote{Details on the design of distribution shift detection pipeline and features extracted are available at \ref{sec:appendix_detailsdistributionshift}}. The model is trained on 5-fold cross-validation splits with the final attack score for each dataset taken as an average of ones obtained across folds. We assume that if both members and non-members come from the same distribution, i.e. the assumption of i.i.d. samples is valid, then this type of attacks should fail, showing AUC-ROC around 50\%. Otherwise, if data collection method was biased (e.g. due to temporal differences, different data generation processes or other factors), these baseline attacks should serve as a lower bound for the proposed membership inference approaches.

We evaluated recently proposed MIA benchmarks in text \cite{fu2024wikimia24, hallinan2025wikimiahard} and image \cite{li2024vlmmembership} modalities using the proposed method, and found that most of them suffer from severe distribution shifts, making them hardly useful to evaluate MIAs, with only LAION-MI \cite{dubiński2023realistic} being mostly unaffected. See ~\autoref{tab:distributionshift}. 
Thus, in order to ensure credible results, we aim to use random splits of recently open-sourced multimodal datasets for Russian language \cite{mera} in our further experiments.
Although we are unaware of any common MIA benchmarks for audio or video data, we release both image and audio pipelines and encourage the community to use them prior to the release of new MIA datasets.

\subsection{Methodology}
\label{sec:method}
Membership inference attacks (MIAs) against LLMs aim to determine whether given a target model $\mathcal{M}$ and a given data point was part of the training dataset used to train the target model. Given a multimodal sample $ x=(t, s) $ from the dataset $ D \sim \mathcal{P}(\mathcal{T} \times \mathcal{S}) $ where $ s \in \mathcal{S} $ is some modality (image/video/audio), $ t \in \mathcal{T} $ is the text, estimate $ \mathbb{P} (x \in D|\mathcal{M}) $, probability that a target model was trained on $x$.

In accordance with the original article \cite{smia}, we divided the training algorithm into the following subsequent steps with some modifications: 
\begin{enumerate}[nosep]
    \item Neighbor generation
    \item Embedding generation
    \item Loss computation
    \item Training the attack model
\end{enumerate}

\subsubsection{Neighbor and embedding generation}
For each original data point $(t, s)$ we generate $K=24$ perturbed "neighbors" $(t^k_\prime, s^k_\prime)$. We apply four different perturbation techniques:
\begin{enumerate}[nosep]
    \item Random masking and predicting the masks with Fred-T5 model~\footnote{\href{https://huggingface.co/ai-forever/FRED-T5-1.7B}{ai-forever/FRED-T5-1.7B}, \cite{zmitrovich2023family}}
    \item Deletion of random tokens
    \item Duplication
    \item Swapping of random tokens
\end{enumerate}
to the text $t$ with each technique applied 6 times. 
Although, in our experiments we fix $s = s^k_\prime, \: \forall s \in D $, so the modality data remains unchanged, the pipeline can be modified to support neighbors from different modalities as well.

Then for each original text $t$ and its neighbors $t^k_\prime$ we extract their text embeddings using a fixed encoder:
$$e=\mathcal{E}(t), \quad e_{k}^{\prime} = \mathcal{E}(t_k^{\prime})$$
where $\mathcal{E}$ is \texttt{intfloat/e5-mistral-7b-instruct}~\footnote{\href{https://huggingface.co/intfloat/e5-mistral-7b-instruct}{intfloat/e5-mistral-7b-instruct} in our experiments. It used to be SoTA on the MTEB benchmark \cite{muennighoff2022mteb} at the time of the model experiments}.

\subsubsection{Loss computation}
We compute the multimodal loss for both models $\mathcal{M}$ and $\mathcal{M}_{leak}$ on both the original and neighbor data points:
$$\mathcal{L} = \mathcal{L}(\mathcal{M}, t, s), \quad \mathcal{L}_k^{\prime} = \mathcal{L}(\mathcal{M}, t^k_{\prime}, s^k_{\prime})$$
Text input $t$ is provided to each model, accompanied by the corresponding modality $s$ (image, video, or audio data in its original, unchanged form). 

\subsubsection{Attack model training}
The core of FiMMIA is a binary neural network classifier trained to distinguish between models that have and have not seen the data. For each neighbor $k$ we create two training examples by computing feature differences\footnote{
Similar ideas has been already explored e.g. in \cite{he2024difficultycalibration} where the authors explored both utilizing shadow models and perturbed datasets as calibration data, and found that they are, to a large degree, interchangeable. The idea of using embedding differences as a proxy for difficulty calibration serves as another intuition for our method.
}:
$$\Delta \mathcal{L} = \mathcal{L} - \mathcal{L}^k_{\prime}, \quad \Delta e = e - e^{k}_{\prime}$$

These feature vectors are paired with labels $y \in \{0, 1\}$ indicating whether the losses came from $\mathcal{M}$ (non-leaked) or $\mathcal{M}_{leak}$ (leaked). However, absolute values of these statistics may vary across datasets and models. To make the system more stable, we apply the z-score normalization technique \cite{StandardScore}. During the training phase, we calculate the mean $\mu$ and standard deviation $\sigma$ for the models loss differences $\Delta \mathcal{L}$ using the evaluation data. 
$$\Delta \mathcal{L}_{norm} = \frac{\Delta \mathcal{L}-\mu}{\sigma} $$.

This process yields random batch training triplets $(\Delta \mathcal{L}_{norm}, \Delta e, y)$ per original data point. The FiMMIA detector, $f_{FiMMIA}$ is trained to predict the probability $p=f_{FiMMIA}(\Delta \mathcal{L}_{norm}, \Delta e)$ that the input features originate from a model that has been trained on the target data. We provide the details of the architecture for FiMMIA model in ~\autoref{sec:appendix_nnarch} and the hyperparameters for training the FiMMIA model in ~\autoref{sec:appendix_hyperparameters}. 

It should be noted, that although we suppose a grey-box access to the MLLM in our experiments, i.e. an attacker has full access to the model's logprobs for loss computation, our setup can be extended to the black-box scenario in presence of compatible APIs, with e.g. only top-k logprobs being released, using approaches from \cite{finlayson2024logits, bao2025glimpseenablingwhiteboxmethods}. We plan to implement such functionality in future releases.

\subsubsection{Inference}
To infer if a target model $\mathcal{M}'$ has been trained on a specific data point $(t, s)$, we compute the loss and embedding differences for this model. We then compute the leakage score $A$ for the data point by taking the average probability output by the detector over all $K$ neighbors:
$$A(t, m) = \frac{1}{K}\sum_{k=1}^{K}f_{FiMMIA}({\Delta \mathcal{L}^k_{norm}}, \Delta e^k)$$

\section{Experiment setup}
\label{sec:experiment}

\subsection{Data}
\label{sec:data}
We evaluate our method on the MERA benchmark \cite{mera}, which comprises 18 audio, video, and image datasets. All tasks in the benchmark are multimodal, taking both a modality input and an instruction, and requiring a text output in a constrained format (e.g., multiple-choice or short-answer).
For training phase we fine-tune MLLM $\mathcal{M}_{leak}$ on each modality separately. Each sample in the training data for the MLLM can be represented as $ x = (s, q, a)$, a concatenation of the question and the answer as the textual part $t$, along with the multimodal input $s$ (image, video, or audio). In order to ensure credible evaluation of  FiMMIA model we split each dataset into train and test parts randomly. The size of the test part is $10\%$ of original dataset. Normalization parameters $\mu_{D,\mathcal{M}}$ and $\sigma_{D,\mathcal{M}}$ are calculated from the train part of each of the splitted datasets for each model.



\subsection{Models}
We evaluate 9 publicly available multimodal models from the most trending model families on HuggingFace, varying in size from 3B to 12B parameters. See Appendix \ref{sec:appendix_models_details} for detailed model descriptions.

\subsection{Cross-lingual transfer}
\label{sec:cross_lingual_transfer}
This section presents our experimental evaluation, extending the pipeline to English image datasets and models. Following the paper by \cite{song-etal-2025-text}, our analysis leverages two multi-choice datasets: ScienceQA \cite{Lu2022LearnTE} and MMStar \cite{Chen2024a}, along with caption dataset: COCO-Caption2017 \cite{Lin2015coco}. We randomly selected 2000 samples from ScienceQA’s test set, respectively, with 1000 samples from the other datasets. We select Qwen2.5-VL-3B-Instruct as a target fine-tuned MLLM and train FiMMIA as described in section ~\autoref{sec:method} only on MERA benchmark \cite{mera} without fine-tuning or using any English data. We evaluate 4 publicly available multimodal models similar to the paper \cite{song-etal-2025-text} that presents MM-DETECT method (see ~\autoref{tab:en_models_image} for model descriptions). That method calculates $\Delta$ score for the dataset and if $\Delta < 0$, dataset leakage is presumed. In order to make a comparison with this method we calculate \% of leaked samples from the dataset, guided by our pipeline. 

\section{Results}
\label{sec:results}
We report AUC-ROC for binary classification (leaked vs. clean) as shown in Tables \ref{tab:FiMMIA_results_image}, \ref{tab:FiMMIA_results_audio}, \ref{tab:FiMMIA_results_video}. Also we report TPR with low FPR in Tables \ref{tab:image_tpr_res}, \ref{tab:video_tpr_res}, \ref{tab:audio_tpr_res} .In order to evaluate the transferability of the trained attack model we also report scores when the origin and test models differ. The $\mathcal{M}_\texttt{origin}$ is the model used to train FiMMIA, while $\mathcal{M}_{\texttt{test}}$ is the model whose losses are used to test FiMMIA (predict whether the data sample was used to train $\mathcal{M}_{\texttt{test}}$ or not).

\begin{table}[t]
\scriptsize
\centering
 \begin{tabular}{l l c}
\toprule[1pt]
$\mathcal{M}_{\texttt{origin}}$ & $\mathcal{M}_{\texttt{test}}$ & \textbf{AUC-ROC} \\
\midrule[1pt]
Qwen2.5-VL-3B-Instruct & Qwen2.5-VL-3B-Instruct & \textbf{96.2} \\
Qwen2.5-VL-3B-Instruct & Qwen2-VL-7B-Instruct & 86.0 \\
Qwen2.5-VL-3B-Instruct & Qwen2.5-VL-7B-Instruct & 88.0 \\
Qwen2.5-VL-3B-Instruct & Llama3-llava-next-8b-hf & \textit{90.2} \\
Qwen2.5-VL-3B-Instruct & Gemma-3-4b-it & 65.8 \\
Qwen2.5-VL-3B-Instruct & Gemma-3-12b-it & 67.9 \\
\midrule
Qwen2-VL-7B-Instruct & Qwen2.5-VL-3B-Instruct & 78.0 \\
Qwen2-VL-7B-Instruct & Qwen2-VL-7B-Instruct & \textbf{96.2} \\
Qwen2-VL-7B-Instruct & Qwen2.5-VL-7B-Instruct & \textit{80.5} \\
Qwen2-VL-7B-Instruct & Llama3-llava-next-8b-hf & 78.0 \\
Qwen2-VL-7B-Instruct & Gemma-3-4b-it & 77.7 \\
Qwen2-VL-7B-Instruct & Gemma-3-12b-it & 73.7 \\
\midrule
Qwen2.5-VL-7B-Instruct & Qwen2.5-VL-3B-Instruct & 92.8 \\
Qwen2.5-VL-7B-Instruct & Qwen2-VL-7B-Instruct & 93.1 \\
Qwen2.5-VL-7B-Instruct & Qwen2.5-VL-7B-Instruct & \textbf{98.1} \\
Qwen2.5-VL-7B-Instruct & Llama3-llava-next-8b-hf & \textit{95.8} \\
Qwen2.5-VL-7B-Instruct & Gemma-3-4b-it & 95.4 \\
Qwen2.5-VL-7B-Instruct & Gemma-3-12b-it & 94.5 \\
\midrule
Llama3-llava-next-8b-hf & Qwen2.5-VL-3B-Instruct & 94.6 \\
Llama3-llava-next-8b-hf & Qwen2-VL-7B-Instruct & 90.0 \\
Llama3-llava-next-8b-hf & Qwen2.5-VL-7B-Instruct & 96.6 \\
Llama3-llava-next-8b-hf & Llama3-llava-next-8b-hf & \textit{97.7} \\
Llama3-llava-next-8b-hf & Gemma-3-4b-it & 99.1 \\
Llama3-llava-next-8b-hf & Gemma-3-12b-it & \textbf{99.5} \\
\midrule
Gemma-3-4b-it & Qwen2.5-VL-3B-Instruct & 76.0 \\
Gemma-3-4b-it & Qwen2-VL-7B-Instruct & 71.5 \\
Gemma-3-4b-it & Qwen2.5-VL-7B-Instruct & 85.2 \\
Gemma-3-4b-it & Llama3-llava-next-8b-hf & 86.5 \\
Gemma-3-4b-it & Gemma-3-4b-it & \textbf{99.4} \\
Gemma-3-4b-it & Gemma-3-12b-it & \textit{98.7} \\
\midrule
Gemma-3-12b-it & Qwen2.5-VL-3B-Instruct & 84.1 \\
Gemma-3-12b-it & Qwen2-VL-7B-Instruct & 81.3 \\
Gemma-3-12b-it & Qwen2.5-VL-7B-Instruct & 91.2 \\
Gemma-3-12b-it & Llama3-llava-next-8b-hf & 93.3 \\
Gemma-3-12b-it & Gemma-3-4b-it & \textit{99.4} \\
Gemma-3-12b-it & Gemma-3-12b-it & \textbf{99.7} \\
\bottomrule[1pt]
\end{tabular}
\caption{AUC-ROC FiMMIA performance metrics for various evaluated \textbf{Image} MLLMs.}
\label{tab:FiMMIA_results_image}
\end{table}

\begin{table}[!ht]
\scriptsize
\centering
\begin{tabular}{l l c}
\toprule[1pt]
$\mathcal{M}_{\texttt{origin}}$ & $\mathcal{M}_{\texttt{test}}$ & \textbf{AUC-ROC} \\
\midrule[1pt]
Qwen2.5-VL-3B-Instruct & Qwen2.5-VL-3B-Instruct & \textit{95.9} \\
Qwen2.5-VL-3B-Instruct & Qwen2.5-VL-7B-Instruct & \textbf{99.5} \\
Qwen2.5-VL-3B-Instruct & LLaVA-NeXT-Video & 91.7 \\
Qwen2.5-VL-3B-Instruct & LLaVA-NeXT-Video-DPO & 91.2 \\
\midrule
Qwen2.5-VL-7B-Instruct & Qwen2.5-VL-3B-Instruct & \textit{98.7} \\
Qwen2.5-VL-7B-Instruct & Qwen2.5-VL-7B-Instruct & \textbf{100.0} \\
Qwen2.5-VL-7B-Instruct & LLaVA-NeXT-Video & 96.5 \\
Qwen2.5-VL-7B-Instruct & LLaVA-NeXT-Video-DPO & 95.7 \\
\midrule
LLaVA-NeXT-Video & Qwen2.5-VL-3B-Instruct & 63.7 \\
LLaVA-NeXT-Video & Qwen2.5-VL-7B-Instruct & \textit{71.5} \\
LLaVA-NeXT-Video & LLaVA-NeXT-Video & \textbf{100.0} \\
LLaVA-NeXT-Video & LLaVA-NeXT-Video-DPO & \textbf{100.0} \\
\midrule
LLaVA-NeXT-Video-DPO & Qwen2.5-VL-3B-Instruct & 53.6 \\
LLaVA-NeXT-Video-DPO & Qwen2.5-VL-7B-Instruct & \textit{56.2} \\
LLaVA-NeXT-Video-DPO & LLaVA-NeXT-Video & \textbf{100.0} \\
LLaVA-NeXT-Video-DPO & LLaVA-NeXT-Video-DPO & \textbf{100.0} \\
\bottomrule[1pt]
\end{tabular}
\caption{AUC-ROC FiMMIA performance metrics for various evaluated \textbf{Video} MLLMs.}
\label{tab:FiMMIA_results_video}
\end{table}

\begin{table}[!ht]
\scriptsize
\centering
\begin{tabular}{l l c}
\toprule[1pt]
$\mathcal{M}_{\texttt{origin}}$ & $\mathcal{M}_{\texttt{test}}$ & \textbf{AUC-ROC} \\
\midrule[1pt]
Qwen2-Audio-7B-Instruct & Qwen2-Audio-7B-Instruct & \textbf{87.7} \\
Qwen2-Audio-7B-Instruct & Qwen-Audio-Chat & \textit{76.0} \\
\midrule
Qwen-Audio-Chat & Qwen2-Audio-7B-Instruct & \textit{61.3} \\
Qwen-Audio-Chat & Qwen-Audio-Chat & \textbf{100.0} \\
\bottomrule[1pt]
\end{tabular}
\caption{AUC-ROC FiMMIA performance metrics for various evaluated \textbf{Audio} MLLMs.}
\label{tab:FiMMIA_results_audio}
\end{table}

Overall, the results of the FiMMIA detection capabilities are presented in \autoref{tab:FiMMIA_results_overall}.
All models show significant success within their own family; however, the success of the attack may decrease when testing on a model from a different family. Nevertheless, the metric score for each experiment exceeds 65.0, which indicates the promising transferability of the proposed method. Moreover, average metrics for each modality are quite high, ranging from 80 to 90\% AUC-ROC.

\begin{table}[!ht]
\small
\centering
\begin{tabular}{lc}
\toprule
\textbf{Modality} & \textbf{AUC-ROC} \\
\midrule
Image & 88.658 \\ 
Video & 88.388 \\ 
Audio & 81.250 \\ 
\bottomrule
\end{tabular}
\caption{Average AUC-ROC of FiMMIA per modality. Averaging over the models used for training and evaluating FiMMIA.}
\label{tab:FiMMIA_results_overall}
\end{table}

Evaluations on the transferability of the model to a different language inputs are presented in ~\autoref{tab:results_language_transfer}. The results indicate that our method is almost entirely in agreement with those presented in the paper \cite{song-etal-2025-text}. If $\Delta<0$ the amount of samples predicted by FiMMIA as leaked is more than $0.1$ in most cases, which corresponds to at least $10\%$ of the dataset. However, if the task allows, we suggest to train FiMMIA for particular dataset and language from scratch to obtain more accurate and reliable results.

\begin{table}[!ht]
\centering
\tiny
\begin{tabular}{l l l c}
\toprule
\textbf{Dataset} & \textbf{Model} & \textbf{FiMMIA} & \textbf{MM-DETECT $\Delta$} \\
\midrule
\multirow{4}{*}{COCO} & Phi-3-vision-128k-instruct & 0.00 & 0.5 \\
 & Qwen-VL-Chat & 0.00 & -1.9 \\
 & LLaVA-1.5-7B  & 0.58  & -0.6 \\
 & fuyu-8b & 0.22 & 1.0 \\
\midrule
\multirow{4}{*}{MMStar} & Phi-3-vision-128k-instruct & 0.06 & 3.2 \\
 & Qwen-VL-Chat & 0.00 & 3.3 \\
 & LLaVA-1.5-7B & 0.13 & 2.8 \\
 & fuyu-8b & 0.011 & -1.2 \\
\midrule
\multirow{4}{*}{ScienceQA} & Phi-3-vision-128k-instruct & 0.10 & 0.7 \\
 & Qwen-VL-Chat & 0.00 & 0.1 \\
 & LLaVA-1.5-7B & 0.21 & 1.3 \\
 & fuyu-8b & 0.19 & -0.5 \\
\bottomrule
\end{tabular}
\caption{Comparison FiMMIA \% leakage samples detected of MLLMs on English datasets with MM-DETECT score for image modality.}
\label{tab:results_language_transfer}
\end{table}
\section{Conclusion}
\label{sec:conclusion}
This paper introduces FiMMIA, a novel framework that leverages input semantics and strategic perturbations to train a highly effective neural network for data leakage detection in MLLMs. Our key contribution is a language-agnostic system capable of training robust leakage detection models for any dataset. Designed for extensibility, the framework natively supports neighbor generation across multiple modalities paving the way for future research. 

\section*{Limitations}
\label{sec:limit}

\paragraph{Scope of the Method} When training FiMMIA, we only target a fine-tuning scenario for the MLLM using a low-rank adapter. The results for pretraining and full fine-tuning may be  different due to the capacity scaling laws \cite{morris2025much}, and other factors. We leave these evaluations for further work.

\paragraph{Determinism and Reproducibility} Even our fine-tuned models' losses are subject to stochasticity, as the entire hardware–software stack affects inference: GPU model, drivers/CUDA/cuDNN, PyTorch, vLLM/transformers (and commit hashes), flash-attention kernels, tokenizers/checkpoints, precision/quantization, and batching -- some of which are non-deterministic or can vary between environments. However, in general, the variance that these factors contribute to evaluation metrics is not substantial.

\paragraph{Speed and Computational Complexity} In our experiments the inference process took appx. 10 hours on a single GPU for one dataset. Generally, the time complexity of our algorithm scales as $ \mathcal{O}(|D|N(M+E+G))$, where $|D|$ is the number of samples in the dataset, $N$ is the number of neighbors, and $M,E,G$ are time complexities of the target, embedding and neighbor generation models.

\paragraph{Model Assumption Dependencies} The method relies on per-sample loss access (a gray-box assumption) and depends on an external model for generating embeddings. The applicability of the method in a strict black-box setting, where such access is unavailable, is not addressed in this work, despite the existence of relevant prior research.


\section*{Ethical consideration}

\paragraph{Use of Public Data}
All experiments and evaluations in this study rely exclusively on openly accessible public datasets. No proprietary, confidential, or otherwise sensitive information was involved. This choice supports transparency, facilitates independent verification, and avoids any infringement on data-privacy protections.

\paragraph{Defensive and Constructive Purpose}
Our work reconceptualizes membership-inference analysis as a diagnostic and privacy-protecting tool rather than a privacy-threat vector. The method is designed to:
\begin{itemize}[nosep]
    \item By identifying cases in which benchmark samples have been inadvertently memorized during training, the approach helps prevent benchmark saturation and dataset contamination, thereby supporting fair and meaningful model comparison.
    \item The technique offers researchers a practical mechanism for auditing training pipelines to ensure that performance improvements stem from genuine advances rather than overfitting to widely used evaluation sets.
    \item As competitive leaderboard dynamics can unintentionally encourage data leakage and undermine the long-term value of public benchmarks, our framework contributes to more resilient evaluation standards that promote steady, reliable scientific progress.
\end{itemize}

\section*{Acknowledgments}
The authors would like to express their sincere gratitude to Dmitry Gorbetsky, Yaroslav Grebnyak, and Artem Chervyakov for their valuable contributions and support in this work.

\bibliography{custom}

\appendix

\section{Appendix}
\label{sec:appendix}
\subsection{Attack model neural network architecture}
\label{sec:appendix_nnarch}
The detailed architecture of the FiMMIA is provided below.
\begin{enumerate}
    \item \textbf{Input Data:}
    \begin{itemize}
        \item \texttt{loss\_input}: A tensor fed into the \texttt{loss\_component}.
        \item \texttt{embedding\_input}: A tensor fed into the \texttt{embedding\_component}.
    \end{itemize}

    \item \textbf{\texttt{loss\_component}:}
    \begin{itemize}
        \item A Linear layer: 1 input feature $\rightarrow$ \texttt{projection\_size} output features.
        \item Dropout(0.2) and  ReLU \cite{Nair2010} activation.
    \end{itemize}

    \item \textbf{\texttt{embedding\_component}:}
    \begin{itemize}
        \item A Linear layer: \texttt{embedding\_size} $\rightarrow$ \texttt{embedding\_size // 2}.
        \item Dropout(0.2) and  ReLU \cite{Nair2010} activation.
        \item A Linear layer: \texttt{embedding\_size // 2} $\rightarrow$ 512.
        \item Dropout(0.2) and  ReLU \cite{Nair2010} activation.
    \end{itemize}

    \item \textbf{Concatenation (\texttt{torch.hstack}):}
    \begin{itemize}
        \item The outputs from the \texttt{loss\_component} (\texttt{projection\_size}) and the \texttt{embedding\_component}(512) are concatenated into a single vector of size \texttt{2 * projection\_size}.
    \end{itemize}

    \item \textbf{\texttt{attack\_encoding}:}
    \begin{itemize}
        \item A series of 6 fully connected Linear layers with Dropout(0.2) and  ReLU \cite{Nair2010} activations between them: \texttt{2 * projection\_size} $\rightarrow$ 512 $\rightarrow$ 256 $\rightarrow$ 128 $\rightarrow$ 64 $\rightarrow$ 32.
        \item The final Linear layer: 32 $\rightarrow$ 2 (output logits for classification).
        \item A final  ReLU \cite{Nair2010} activation after the last layer.
    \end{itemize}

    \item \textbf{Output:}
    \begin{itemize}
        \item The model returns the logits (size 2).
        \item If labels are provided, it also calculates and returns the cross-entropy loss \cite{Mao2023}.
    \end{itemize}
\end{enumerate}

\subsection{Attack model hyperparameters}
\label{sec:appendix_hyperparameters}

To construct the neighbor datasets, we generate $k = 24$ neighbors for each data point. We employ the adafactor optimizer \cite{adafactor} to train the network on our training data over 10 epochs. The batch size is set to 64, meaning each batch contains random triplets. For experiments, we use a learning rate of $2 \times 10^{-6}$.

\subsection{Models Details}
\label{sec:appendix_models_details}

~\autoref{tab:baselines_models_image} contains information about multimodal LLMs used for the experiments. As the number of MLLMs trained with a focus on russian is limited, we evaluate our method using known open-source models. Although it may contribute to higher ROC-AUC scores we observe in our experiments due to the models being adapted to vastly new domain, it also helps us alleviate possible effects related to the possibility of our evaluation datasets' traces being already present in models' training data.

\begin{table*}[t]
    \setlength{\tabcolsep}{3pt}
    \centering
    \small
    \begin{tabularx}{\textwidth}{%
    @{}c%
    p{0.22\linewidth}%
    r%
    r%
    p{0.30\linewidth}%
    l%
    @{}l@{}}
        \toprule
        & \textbf{Model} & \textbf{Parameters} & \textbf{Context length} & \textbf{Hugging Face Hub link} & \textbf{Citation} \\
        \midrule
        & Qwen2-VL-7B-Instruct & 7B & 32K & \href{https://huggingface.co/Qwen/Qwen2-VL-7B-Instruct}{Qwen/Qwen2-VL-7B-Instruct} & {\citet{wang2024qwen2vlenhancingvisionlanguagemodels}}\\
        \cmidrule{2-6}
        & Qwen2.5-VL-3B-Instruct & 3B & 128K & \href{https://huggingface.co/Qwen/Qwen2.5-VL-3B-Instruct}{Qwen/Qwen2.5-VL-3B-Instruct} & \multirow{2}{*}{\citet{bai2025qwen25vltechnicalreport}}\\
        & Qwen2.5-VL-7B-Instruct & 7B & 128K & \href{https://huggingface.co/Qwen/Qwen2.5-VL-7B-Instruct}{Qwen/Qwen2.5-VL-7B-Instruct} & \\
        \cmidrule{2-6}
        & gemma-3-4b-it & 4B & 128K & \href{https://huggingface.co/google/gemma-3-4b-it}{google/gemma-3-4b-it} & \multirow{2}{*}{\citet{gemmateam2025gemma3technicalreport}}\\
        & gemma-3-12b-it & 12B & 128K & \href{https://huggingface.co/google/gemma-3-12b-it}{google/gemma-3-12b-it} & \\
        \cmidrule{2-6}
        & llama3-llava-next-8b-hf & 8B & 128K & \href{https://huggingface.co/llava-hf/llama3-llava-next-8b-hf}{llava-hf/llama3-llava-next-8b-hf} & \citet{li2024llavanext}\\
        \cmidrule{2-6}
        & LLaVA-NeXT-Video & 7B & 4K & \href{https://huggingface.co/llava-hf/LLaVA-NeXT-Video-7B-hf}{llava-hf/LLaVA-NeXT-Video-7B-hf} & \multirow{2}{*}{\citet{liu2024llavanext}}\\
        & LLaVA-NeXT-Video-DPO & 7B & 4K & \href{https://huggingface.co/llava-hf/LLaVA-NeXT-Video-7B-DPO-hf}{llava-hf/LLaVA-NeXT-Video-7B-DPO-hf} & \\
        \cmidrule{2-6}
        & Qwen2-Audio-7B-Instruct & 7B & 32K & \href{https://huggingface.co/Qwen/Qwen2-Audio-7B-Instruct}{Qwen/Qwen2-Audio-7B-Instruct} & \citet{chu2024qwen2audiotechnicalreport}\\
        \cmidrule{2-6}
        & Qwen/Qwen-Audio-Chat & 7B & 32K & \href{https://huggingface.co/Qwen/Qwen-Audio-Chat}{Qwen/Qwen-Audio-Chat} & \citet{qwen_audio}\\
        \bottomrule
    \end{tabularx}
    \caption{General information about used multimodal LLMS for experiments.}
    \label{tab:baselines_models_image}
\end{table*}

\subsection{English Models Details}
\label{sec:appendix_en_models_details}

~\autoref{tab:en_models_image} contains information about multimodal LLMs used for the language transfer experiments. All models are selected from the following paper \cite{song-etal-2025-text}.

\subsection{TPR at low FPR (FPR=5\%) results}
\label{sec:appendix_tprfpr_res}
Here we report the True Positive Rate (TPR) at a low False Positive Rate (FPR), which measures the detection rate at a meaningful threshold. The modality of image is presented in~\autoref{tab:image_tpr_res}, the video in~\autoref{tab:video_tpr_res} and the audio accordingly in~\autoref{tab:audio_tpr_res}.

\begin{table*}[t]
    \setlength{\tabcolsep}{3pt}
    \centering
    \small
    \begin{tabularx}{\textwidth}{%
    @{}c%
    p{0.22\linewidth}%
    r%
    r%
    p{0.30\linewidth}%
    l%
    @{}l@{}}
        \toprule
        & \textbf{Model} & \textbf{Parameters} & \textbf{Context length} & \textbf{Hugging Face Hub link} & \textbf{Citation} \\
        \midrule
        & Phi-3-vision-128k-instruct & 8B & 128K & \href{https://huggingface.co/microsoft/Phi-3-vision-128k-instruct}{microsoft/Phi-3-vision-128k-instruct} & \cite{abdin2024phi3} \\
         \midrule
        & LLaVA-1.5-7B & 7B & 16K & \href{https://huggingface.co/llava-hf/llava-1.5-7b-hf}{llava-hf/llava-1.5-7b-hf} & \cite{liu2024improvedbaselinesvisualinstruction} \\
        \midrule
        & Qwen-VL-Chat & 7B & 8K& \href{https://huggingface.co/Qwen/Qwen-VL-Chat}{Qwen-VL-Chat} & \cite{Bai2023QwenVA} \\
         \midrule
        &  fuyu-8b\footnote{\url{https://www.adept.ai/blog/fuyu-8b}} & 8B & 16K & \href{https://huggingface.co/adept/fuyu-8b}{adept/fuyu-8b} &  \\
        \bottomrule
    \end{tabularx}
    \caption{General information about used multimodal LLMS used for the language transfer experiments.}
    \label{tab:en_models_image}
\end{table*}

\begin{table}[t]
\tiny
\centering
\setlength{\tabcolsep}{3pt} 
\begin{tabular}{lllll}
\toprule[1pt]
$\mathcal{M}_{\texttt{origin}}$ & $\mathcal{M}_{\texttt{test}}$ & \textbf{AUC-ROC}  & \textbf{TPR} \\
\midrule[1pt]
Qwen2.5-VL-3B-Instruct & Qwen2.5-VL-3B-Instruct & 95.9 & 85.8 \\
Qwen2.5-VL-3B-Instruct & Qwen2.5-VL-7B-Instruct & 99.5 & 98.4 \\
Qwen2.5-VL-3B-Instruct & LLaVA-NeXT-Video & 91.7 & 52.9 \\
Qwen2.5-VL-3B-Instruct & LLaVA-NeXT-Video-DPO & 91.2 & 62.9 \\
\midrule
Qwen2.5-VL-7B-Instruct & Qwen2.5-VL-3B-Instruct & 98.7 & 95.4 \\
Qwen2.5-VL-7B-Instruct & Qwen2.5-VL-7B-Instruct & 100.0 & 100.0 \\
Qwen2.5-VL-7B-Instruct & LLaVA-NeXT-Video & 96.5 & 80.8 \\
Qwen2.5-VL-7B-Instruct & LLaVA-NeXT-Video-7B-DPO & 95.7 & 82.1 \\
\midrule
LLaVA-NeXT-Video & Qwen2.5-VL-3B-Instruct & 63.7 & 6.0 \\
LLaVA-NeXT-Video & Qwen2.5-VL-7B-Instruct & 71.5 & 70.0 \\
LLaVA-NeXT-Video & LLaVA-NeXT-Video-7B & 100.0 & 100.0 \\
LLaVA-NeXT-Video & LLaVA-NeXT-Video-7B-DPO & 100.0 & 100.0 \\
\midrule
LLaVA-NeXT-Video-7B-DPO & Qwen2.5-VL-3B-Instruct & 53.6 & 60.0 \\
LLaVA-NeXT-Video-7B-DPO & Qwen2.5-VL-7B-Instruct & 56.2 & 43.0 \\
LLaVA-NeXT-Video-7B-DPO & LLaVA-NeXT-Video-7B & 100.0 & 100.0 \\
LLaVA-NeXT-Video-7B-DPO & LLaVA-NeXT-Video-7B-DPO & 100.0 & 100.0 \\
\bottomrule
\end{tabular}
\caption{AUC-ROC and TPR at low FPR (FPR=5\%) FiMMIA performance metrics for various evaluated \textbf{Video} MLLMs.}
\label{tab:video_tpr_res}
\end{table}

\begin{table}[t]
\tiny
\centering
\begin{tabular}{lllll}
\toprule[1pt]
$\mathcal{M}_{\texttt{origin}}$ & $\mathcal{M}_{\texttt{test}}$ & \textbf{AUC-ROC}  & \textbf{TPR} \\
\midrule[1pt]
Qwen2-Audio-7B-Instruct & Qwen2-Audio-7B-Instruct & 87.7 & 61.9 \\
Qwen2-Audio-7B-Instruct & Qwen-Audio-Chat & 76.0 & 74.5 \\
\midrule
Qwen-Audio-Chat & Qwen2-Audio-7B-Instruct & 61.3 & 62.7 \\
Qwen-Audio-Chat & Qwen-Audio-Chat & 100.0 & 100.0 \\
\bottomrule
\end{tabular}
\caption{AUC-ROC and TPR at low FPR (FPR=5\%) FiMMIA performance metrics for various evaluated \textbf{Audio} MLLMs.}
\label{tab:audio_tpr_res}
\end{table}

\begin{table}[t]
\scriptsize
\centering

\begin{tabular}{lllll}
\toprule[1pt]
$\mathcal{M}_{\texttt{origin}}$ & $\mathcal{M}_{\texttt{test}}$ & \textbf{AUC-ROC}  & \textbf{TPR} \\
\midrule[1pt]
Qwen2.5-VL-3B-Instruct & Qwen2.5-VL-3B-Instruct & 96.2 & 86.1 \\
Qwen2.5-VL-3B-Instruct & Qwen2-VL-7B-Instruct & 86.0 & 39.1 \\
Qwen2.5-VL-3B-Instruct & Qwen2.5-VL-7B-Instruct & 88.0 & 53.0 \\
Qwen2.5-VL-3B-Instruct & llama3-llava-next-8b-hf & 90.2 & 59.9 \\
Qwen2.5-VL-3B-Instruct & gemma-3-4b-it & 65.8 & 6.2 \\
Qwen2.5-VL-3B-Instruct & gemma-3-12b-it & 67.9 & 61.9 \\
\midrule
Qwen2-VL-7B-Instruct & Qwen2.5-VL-3B-Instruct & 78.0 & 16.5 \\
Qwen2-VL-7B-Instruct & Qwen2-VL-7B-Instruct & 96.2 & 85.1 \\
Qwen2-VL-7B-Instruct & Qwen2.5-VL-7B-Instruct & 80.5 & 35.9 \\
Qwen2-VL-7B-Instruct & llama3-llava-next-8b-hf & 78.0 & 30.6 \\
Qwen2-VL-7B-Instruct & gemma-3-4b-it & 77.7 & 7.2 \\
Qwen2-VL-7B-Instruct & gemma-3-12b-it & 73.7 & 67.8 \\
\midrule
Qwen2.5-VL-7B-Instruct & Qwen2.5-VL-3B-Instruct & 92.8 & 73.8 \\
Qwen2.5-VL-7B-Instruct & Qwen2-VL-7B-Instruct & 93.1 & 77.0 \\
Qwen2.5-VL-7B-Instruct & Qwen2.5-VL-7B-Instruct & 98.1 & 94.0 \\
Qwen2.5-VL-7B-Instruct & llama3-llava-next-8b-hf & 95.8 & 83.1 \\
Qwen2.5-VL-7B-Instruct & gemma-3-4b-it & 95.4 & 71.8 \\
Qwen2.5-VL-7B-Instruct & gemma-3-12b-it & 94.5 & 66.1 \\
\midrule
llama3-llava-next-8b-hf & Qwen2.5-VL-3B-Instruct & 94.6 & 78.6 \\
llama3-llava-next-8b-hf & Qwen2-VL-7B-Instruct & 90.0 & 65.7 \\
llama3-llava-next-8b-hf & Qwen2.5-VL-7B-Instruct & 96.6 & 90.9 \\
llama3-llava-next-8b-hf & llama3-llava-next-8b-hf & 97.7 & 93.3 \\
llama3-llava-next-8b-hf & gemma-3-4b-it & 99.1 & 98.2 \\
llama3-llava-next-8b-hf & gemma-3-12b-it & 99.5 & 99.6 \\
\midrule
gemma-3-4b-it & Qwen2.5-VL-3B-Instruct & 76.0 & 20.2 \\
gemma-3-4b-it & Qwen2-VL-7B-Instruct & 71.5 & 19.6 \\
gemma-3-4b-it & Qwen2.5-VL-7B-Instruct & 85.2 & 42.7 \\
gemma-3-4b-it & llama3-llava-next-8b-hf & 86.5 & 41.7 \\
gemma-3-4b-it & gemma-3-4b-it & 99.4 & 98.0 \\
gemma-3-4b-it & gemma-3-12b-it & 98.7 & 92.7 \\
\midrule
gemma-3-12b-it & Qwen2.5-VL-3B-Instruct & 84.1 & 49.4 \\
gemma-3-12b-it & Qwen2-VL-7B-Instruct & 81.3 & 50.0 \\
gemma-3-12b-it & Qwen2.5-VL-7B-Instruct & 91.2 & 74.2 \\
gemma-3-12b-it & llama3-llava-next-8b-hf & 93.3 & 77.2 \\
gemma-3-12b-it & gemma-3-4b-it & 99.4 & 97.6 \\
gemma-3-12b-it & gemma-3-12b-it & 99.7 & 98.4 \\
\bottomrule
\end{tabular}
\caption{AUC-ROC and TPR at low FPR (FPR=5\%) FiMMIA performance metrics for various evaluated \textbf{Image} MLLMs.}
\label{tab:image_tpr_res}
\end{table}

\subsection{Description of the distribution shift detection pipelines}
\label{sec:appendix_detailsdistributionshift}

For the information on the features extracted from image and audio data see \autoref{tab:feature-extraction}.

\begin{table*}[h]
\setlength{\tabcolsep}{3pt}
\centering
\small
\begin{tabular}{p{0.2\textwidth}p{0.35\textwidth}p{0.35\textwidth}}
\hline
\textbf{Feature Type} & \textbf{Image Features} & \textbf{Audio Features} \\
\hline
\textbf{Texture/Pattern} & 
\begin{itemize}
\item Local Binary Patterns (LBP) histogram
\item SIFT Bag of Visual Words (BoVW)
\end{itemize} &
\begin{itemize}
\item MFCCs (mean coefficients)
\item Chroma features (mean)
\item Tonnetz features (mean)
\end{itemize} \\
\hline
\textbf{Spectral/Frequency} & 
\begin{itemize}
\item DCT coefficients (low-frequency)
\end{itemize} &
\begin{itemize}
\item Spectral centroid (mean)
\item Spectral bandwidth (mean)
\item Spectral rolloff (mean)
\end{itemize} \\
\hline
\textbf{Color/Energy} & 
\begin{itemize}
\item HSV histograms (H, S, V channels)
\end{itemize} &
\begin{itemize}
\item RMS energy (mean)
\item Zero-crossing rate (mean)
\end{itemize} \\
\hline
\textbf{Temporal/Rhythmic} &
\begin{itemize}
    \item ---
\end{itemize} &
\begin{itemize}
\item Tempogram features (mean)
\end{itemize} \\
\hline
\end{tabular}
\caption{Statistical Features Extracted for Image and Audio Classification}
\label{tab:feature-extraction}
\end{table*}

\end{document}